\documentclass[10pt,twocolumn,letterpaper]{article}

\usepackage{cvpr}              
\usepackage[accsupp]{axessibility}

\usepackage[dvipsnames]{xcolor}

\definecolor{cvprblue}{rgb}{0.21,0.49,0.74}
\usepackage[pagebackref,breaklinks,colorlinks,citecolor=cvprblue]{hyperref}

\def\paperID{15697} 
\def\confName{CVPR}
\def\confYear{2024}

\title{BodyMAP - Jointly Predicting Body Mesh and 3D Applied Pressure Map for People in Bed}

\author{
Abhishek Tandon\textsuperscript{*}
\qquad
Anujraaj Goyal\textsuperscript{*}
\\
Carnegie Mellon University\\
United States\\
{\tt\small \{abhishektnd, argo\}@cmu.edu}
\and
Henry M. Clever\\
NVIDIA\\
United States\\
{\tt\small hclever@nvidia.com}
\and
Zackory Erickson\\
Carnegie Mellon University\\
United States\\
{\tt\small zackory@cmu.edu}
}

\begin{document}
\twocolumn[{
\renewcommand\twocolumn[1][]{1}
\maketitle
\includegraphics[width=1.0\linewidth]{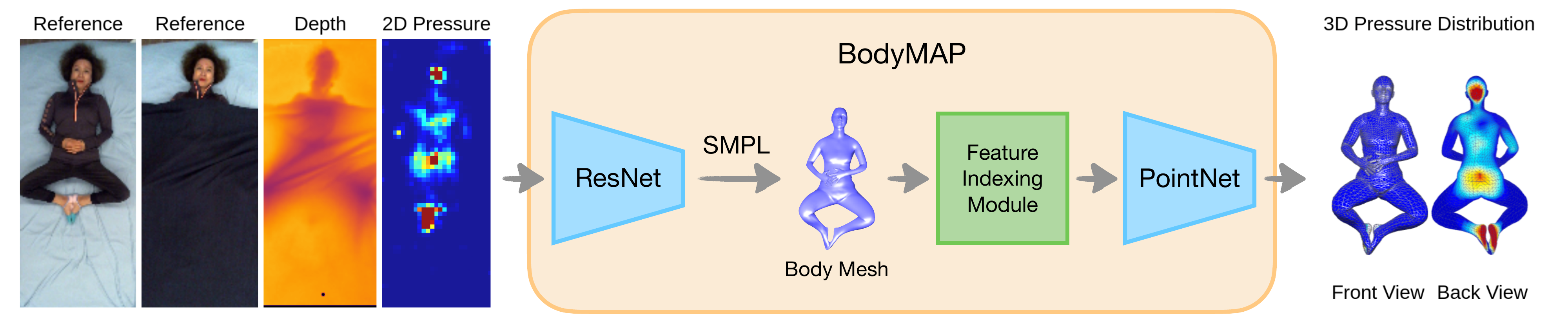}
\captionof{figure}{BodyMAP leverages a depth and pressure image of a person in bed covered by a blanket, to jointly predict the body mesh and a 3D pressure map of pressure distributed along the human body.}
\label{fig:intro}
}]

\let\thefootnote\relax\footnotetext{\textsuperscript{*} Equal contribution.}

\begin{abstract}
\vspace{-2.5mm}
Accurately predicting the 3D human posture and the pressure exerted on the body for people resting in bed, visualized as a body mesh (3D pose \& shape) with a 3D pressure map, holds significant promise for healthcare applications, particularly, in the prevention of pressure ulcers. Current methods focus on singular facets of the problem---predicting only 2D/3D poses, generating 2D pressure images, predicting pressure only for certain body regions instead of the full body, or forming indirect approximations to the 3D pressure map. In contrast, we introduce BodyMAP, which jointly predicts the human body mesh and 3D applied pressure map across the entire human body. Our network leverages multiple visual modalities, incorporating both a depth image of a person in bed and its corresponding 2D pressure image acquired from a pressure-sensing mattress. The 3D pressure map is represented as a pressure value at each mesh vertex and thus allows for precise localization of high-pressure regions on the body. Additionally, we present BodyMAP-WS, a new formulation of pressure prediction in which we implicitly learn pressure in 3D by aligning sensed 2D pressure images with a differentiable 2D projection of the predicted 3D pressure maps. In evaluations with real-world human data, our method outperforms the current state-of-the-art technique by 25\% on both body mesh and 3D applied pressure map prediction tasks for people in bed.
\end{abstract}

\section{Introduction}
\label{sec:intro}

\begin{figure}[t]
\begin{center}
\includegraphics[width=1.0\linewidth]{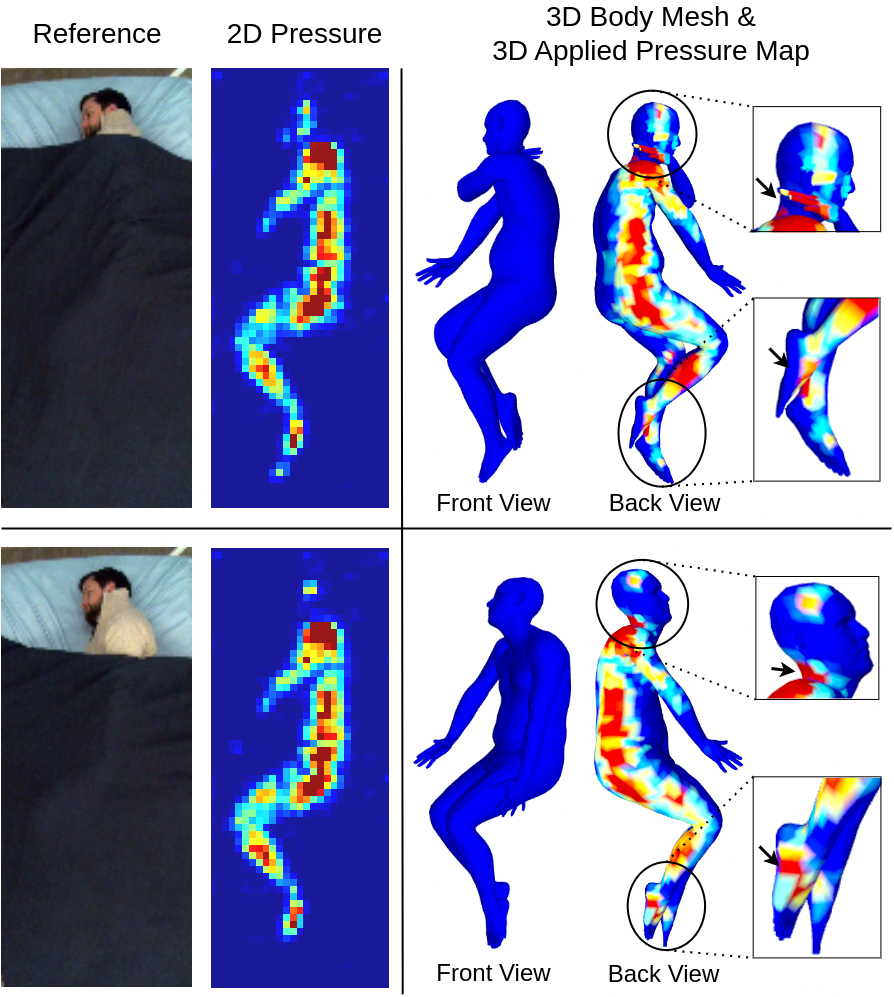}
\end{center}
   \caption{Distinct postures can have similar 2D pressure images. The insets of the 3D pressure map show pressure being applied to different areas demonstrating its use in localizing pressure that is applied on the human body.}
\label{fig:2D3DP}
\end{figure}

With 2.5 million cases annually in the U.S. alone, pressure ulcers remain a pressing concern in healthcare systems worldwide~\cite{berlowitz2011preventing}. Although body repositioning serves as a common preventive measure~\cite{8755997}, concerns persist about whether such pose shifts effectively redistribute applied pressure on the body~\cite{scott2014visual}. Wearable devices~\cite{sen2018new, minteer2020pressure} while aiding in localizing peak pressure on the body, can disrupt normal physical activities and contribute to pressure injuries themselves~\cite{black2010medical, kayser2018prevalence, jackson2019medical}. In response, clinics and long-term care facilities are increasingly looking at pressure sensor systems to aid in pressure ulcer prevention~\cite{8755997, walia2016efficacy, hultin2017pressure}. These systems utilize a pressure sensing array, which is placed beneath the person, and measures the pressure applied by the human body onto the sensor array, producing a 2D pressure image. While informative about pressure applied on the body~\cite{scott2014visual, gunningberg2016facilitating, behrendt2014continuous}, these pressure images have inherent ambiguities due to their 2D depiction of pressure. As illustrated in \cref{fig:2D3DP}, distinct body postures can result in remarkably similar pressure images, failing to correctly convey which body parts are under high pressure. These systems thus currently require a human caregiver to deduce the correct 3D distribution of pressure on the human body, based on the 2D pressure image and 3D body posture. This step not only adds cognitive workload to caregivers but also introduces delays or potential errors in implementing preventive measures. Moreover, individuals in bed are often covered with blankets, making it challenging for nurses or caregivers to accurately identify high-pressure regions on a care recipient's body.

Addressing these challenges, we propose BodyMAP (Fig.~\ref{fig:intro}), a deep-learning model, that jointly estimates a human body mesh (3D pose \& shape) and a 3D pressure distribution on the body mesh for people in bed, covered with blankets. Visualizing the pressure map on the 3D human body mesh, as illustrated in \cref{fig:2D3DP}, precisely pinpoints body regions under peak pressure. Automatic body mesh and 3D pressure map predictions could reduce the need for caregivers to manually infer them, and offer visual insights into pressure redistribution as caregivers reposition a person's body. This real-time feedback on peak body pressure can transform the current `blind' body repositioning into an informed process, enhancing care quality and potentially reducing the occurrence of preventable pressure injuries. Additionally, other domains requiring body mesh and pressure knowledge such as assistive robotics~\cite{clever20183d, wang2022visual}, sports rehabilitation~\cite{5690041}, and eldercare applications~\cite{nishi2017generation} could benefit from accurate pressure prediction on the human body. 

BodyMAP uses a combination of a depth image and the corresponding 2D pressure image, for an individual in bed, as inputs to predict the body mesh and the applied 3D pressure map. A depth camera, placed vertically above the bed, provides a top-down view of the mattress, while a fabric pressure sensing array placed beneath the person, provides the `bottom-up view' of the body. We represent the human body using the SMPL parametric model~\cite{loper2023smpl} and predict the SMPL parameters related to body shape and joint angles. Our method first predicts the SMPL mesh and then leverages the predicted mesh vertices and latent features to estimate a pressure value at each vertex, producing a 3D pressure map. To train our model, we obtain ground truth applied pressure map data by projecting 2D pressure images onto corresponding ground truth body meshes~\cite{clever2022bodypressure}. Our project page containing code, trained models and the 3D pressure ground truth data for SLP~\cite{liu2022simultaneously, liu2019seeing} and BodyPressureSD~\cite{clever2022bodypressure} datasets used in our work are available for research purposes at \href{https://bodymap3d.github.io/}{https://bodymap3d.github.io/}.

Prior methods either predict pressure maps for a subset of body regions~\cite{wang2022visual} or indirectly approximate 3D pressure~\cite{clever2022bodypressure}. In contrast, BodyMAP uses a unified model to jointly predict the human mesh and 3D applied pressure map across the entire body. This joint training approach offers several advantages, including human-interpretable visualizations of pressure on the human body mesh for caregivers, reduced model inference time, and more accurate predictions of body pose and 3D pressure. 

Additionally, we introduce BodyMAP-WS (3D applied pressure prediction without supervision), a variant of our model, designed to implicitly learn the 3D pressure map without direct supervision. BodyMAP-WS constructs a differentiable 2D projection of a predicted 3D pressure map and aligns the 2D projection with the input pressure image. This alignment encourages the model to learn the correct 3D pressure map without relying on ground truth pressure maps on the human body that are difficult to measure in the real world. By removing the need for labels, this technique holds the potential to reduce data collection efforts and allow for training models on unlabeled data.

Our method sets a new state-of-the-art in predicting body mesh (3D pose \& shape) and 3D applied pressure on the human body, surpassing previous works by 25\% on both tasks, as evidenced by our model's performance across various 3D pose, 3D shape, and 3D pressure metrics. In summary, our contributions include the following:
\begin{itemize}
    \item A deep learning method, BodyMAP, that takes as input a depth image and the corresponding 2D pressure image for a person in bed and infers jointly the body mesh and the 3D pressure map of pressure applied on the body. 
    \item An implicit 3D pressure map prediction method, BodyMAP-WS, to support scaling up body pressure prediction models on real-world unlabeled datasets.
\end{itemize}
\section{Related Work}
\label{sec:relatedwork}

\subsection{Pose estimation using deep learning}
Pose estimation is a well-established task in Computer Vision~\cite{agarwal2005recovering, toshev2014deeppose,  tian2023recovering, liu2022recent}. Parametric human models~\cite{loper2023smpl, pavlakos2019expressive} like SMPL simplify the representation of the human body, forming a dense 3D body mesh from only a small set of parameters for body shape and joint angles. 

While supervised methods~\cite{lin2021end, kolotouros2019learning} use 3D ground truth data, recent efforts have explored geometric cues and multi-view consistency to predict body mesh from RGB images~\cite{chen2019unsupervised, kocabas2019self, wang20193d}. Gong~\etal~\cite{gong2022self} use multiple 2D modalities to predict the SMPL body and employ self-correction between the input and 2D projections to train their model. Inspired by this line of research in pose estimation, we developed our method, BodyMAP-WS to learn 3D pressure maps without direct supervision. 

\subsection{In-bed human pose estimation}

In-bed human poses present distinct challenges compared to active poses observed in sports activities. Heavy occlusion caused by blankets and self-occlusion present complexities. Additionally, there is limited body visibility in modalities such as pressure images which fail to capture limbs that are not in contact with the mattress. Datasets like the SLP dataset~\cite{liu2022simultaneously, liu2019seeing} featuring real-world human participants and the synthetic BodyPressureSD dataset~\cite{clever2022bodypressure}, which we leverage in this study, have propelled research in this domain.

Yin~\etal~\cite{yin2022multimodal} introduce Pyramid Fusion, incorporating modalities like RGB, pressure, depth, and IR images. However, their design necessitates multiple model passes to derive the final predicted body mesh. Considering potential clinical deployments and privacy concerns, we move away from using RGB images in our work. Clever~\etal propose methods~\cite{clever2020bodies, clever2022bodypressure} involving multiple models to estimate the body mesh. These approaches only rely on a single modality (pressure image and depth respectively). Our method, BodyMAP, simplifies this process by employing a single pass through a single model to jointly predict both the body mesh and applied 3D pressure map, while utilizing multiple visual modalities as input. Our findings demonstrate that our streamlined approach significantly improves performance over prior methodologies.

\subsection{3D applied pressure map prediction}

Prior techniques~\cite{clever2022bodypressure, liu2023pressure} have explored the task of predicting 2D pressure images from alternative modalities like depth and RGB images respectively, offering a cost-effective means of obtaining information about applied pressure. However, as illustrated in \cref{fig:2D3DP}, 2D pressure images fail to correctly identify the body regions under high pressure.

In the related domain of pain monitoring, 3D pain drawings have emerged to be more effective visualization tools than their 2D counterparts~\cite{ghinea20083, spyridonis20113} in analyzing body pain. In this work, we expand on this concept by directly predicting the 3D applied pressure map onto a human body mesh, overcoming the limitations of 2D pressure images.

Only a few prior methods explore predicting pressure on the human body. Wang~\etal~\cite{wang2022visual} predict pressure applied during cloth interaction, only on specific body regions such as arms. 
Clever~\etal~\cite{clever2022bodypressure} proposed the concept of full-body 3D pressure maps. The authors developed a multi-model pipeline trained in multiple stages, to predict both the body mesh and 2D pressure image from a depth image. Finally, they approximate the 3D pressure map by vertically projecting the 2D pressure image over the estimated 3D human body. We note that errors from each model often compound, affecting the accuracy of 3D pressure map predictions.

In contrast, we devise a unified model architecture, BodyMAP, to estimate the body mesh and pressure map jointly. Our model leverages PointNet~\cite{qi2017pointnet} to predict 3D pressure map at the vertex level across the entire body mesh.

\section{Method}

\begin{figure*}
\begin{center}
\includegraphics[width=1.0\linewidth]{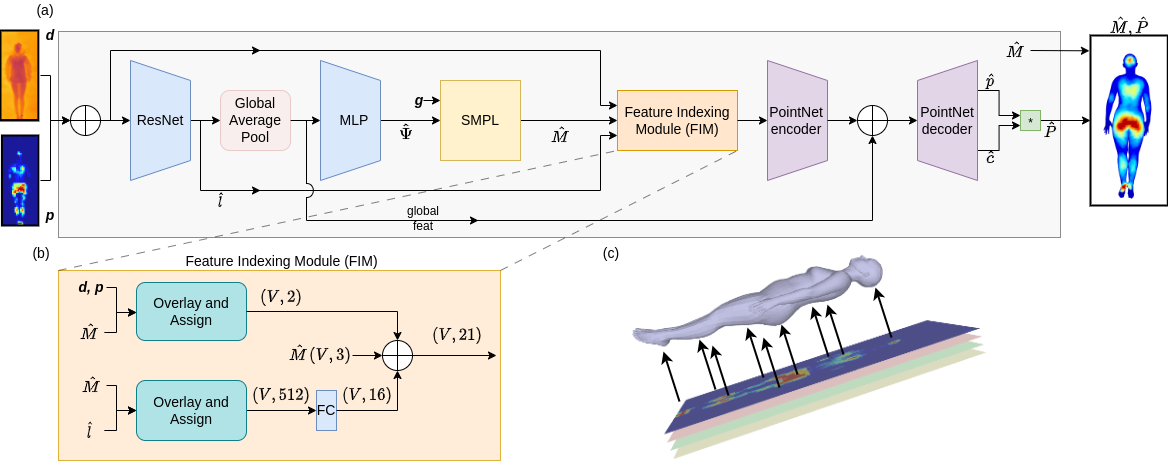}
\end{center}
   \caption{BodyMAP jointly predicts body mesh and 3D applied pressure map for an individual in bed. (a) Model architecture that encodes depth image \textbf{\textit{d}} and 2D pressure image \textbf{\textit{p}} to predict SMPL \cite{loper2023smpl} parameters $\hat{\Psi}$, used to reconstruct the SMPL mesh $\hat{M}$. The Feature Indexing Module (FIM) accumulates features for the mesh vertices from the input images and ResNet features. PointNet predicts the per-vertex pressure \textit{$\hat{p}$} \& per-vertex binary contact \textit{$\hat{c}$} along the human body using the mesh features as input. Finally, the 3D pressure map \textit{$\hat{P}$} is calculated as the product of the per-vertex pressure value and binary contact value. (b) FIM overlays the predicted mesh over the ResNet feature maps (\textit{$\hat{l}$}) and input images by mapping mesh vertex locations to pixel positions and then assigns features to each vertex (V). These are fused along with the mesh vertex locations and used for 3D pressure map prediction. (c) FIM's `overlay and assign' step visualization.}
\label{fig:modelfull}
\end{figure*}

We train a deep-learning model, BodyMAP, to jointly predict the body mesh \textit{$\hat{M}$} (3D pose \& shape), along with the 3D applied pressure map \textit{$\hat{P}$}. Specifically, our model takes the individual's gender \textit{g}, the depth image \textit{d}, and the 2D pressure image \textit{p} as inputs.

The depth image is captured by a depth camera situated above the bed, while the 2D pressure image is generated by a pressure sensing system positioned beneath the individual. This arrangement captures complementary features regarding the body, illustrated in the appendix \cref{fig:smod}, thereby enhancing the context available to the model for accurate predictions of both body mesh and pressure map.

The human body is represented using the SMPL~\cite{loper2023smpl} mesh while the 3D pressure map is represented at the vertex level with a pressure value for each vertex of the human mesh, allowing for peak pressure localization on the body.

We train our models on both the BodyPressureSD dataset~\cite{clever2022bodypressure} of simulated humans in bed and the real-world SLP dataset~\cite{liu2019seeing, liu2022simultaneously}. For both datasets, we have depth and pressure images, aligned with the 3D ground truth mesh for diverse poses (supine, left \& right lateral) and multiple blanket thickness configurations.

\subsection{BodyMAP}

In this section, we detail the architecture of BodyMAP, as illustrated in \cref{fig:modelfull}. The depth and pressure images are resized, concatenated and processed together as image channels by the model. 

\textbf{Body mesh prediction}: The input is first encoded using ResNet18~\cite{he2016deep}. The latent features are then passed through a multi-layer perceptron to predict the SMPL parameters \textit{$\hat{\Psi}$}. These parameters include body shape, joint angles, root-joint translation, and root-joint rotation. The SMPL parameters $\hat{\Psi}$ and gender information \textit{g}, serve as inputs to the SMPL embedding block~\cite{kanazawa2018end}. This block does not contain any learned parameters and outputs a differentiable human body mesh \textit{$\hat{M}$} with vertices \textit{$\hat{V}$}, and 3D joint positions \textit{$\hat{S}$}. 

\textbf{Feature Indexing Module}: We introduce the Feature Indexing Module (FIM), depicted in \cref{fig:modelfull}(b), to accumulate features for each mesh vertex. As mentioned in~\cite{clever2022bodypressure}, the mesh predictions are spatially registered with the corresponding input images. This registration provides pixel locations where each mesh vertex would project onto the input images~\cite{clever2022bodypressure}. FIM assigns features to each mesh vertex from both the input images and the latent ResNet features (before global average pooling) using the mentioned pixel locations (\cref{fig:modelfull}(c)). These features are fused with the vertex locations and are utilized for pressure map prediction. 

\textbf{Pressure map prediction}: To predict a per-vertex pressure map (\textit{$\hat{P}$}), we employ PointNet~\cite{qi2017pointnet} utilizing the latent features formed from FIM for each vertex as input. This establishes a strong correlation between the mesh prediction and pressure prediction, ensuring their consistency with each other. In a manner akin to point-based segmentation architectures~\cite{qi2017pointnet++, qi2017pointnet}, ResNet features (after global average pooling) are fused with PointNet encoder features. This provides the model with an enhanced contextual understanding. The PointNet decoder then predicts a per-vertex binary contact value and pressure value. The per-vertex contact value serves as an indicator of whether each mesh vertex is in contact with the mattress. We use the predicted contact values to further tune the predicted pressure values. Specifically, the 3D pressure map is finally estimated as a product of the binary contact value with the corresponding pressure value for each vertex. 

\subsubsection{Training strategy}

We train the network to jointly predict body mesh and 3D applied pressure map with the following loss: 
\begin{equation} \label{eq:1}
    L = \mathcal{L}_{\text{SMPL}} + \lambda_{1}\mathcal{L}_{\text{v2v}} + \lambda_{2}\mathcal{L}_{\text{P3D}} + \lambda_{3}\mathcal{L}_{\text{contact}}
\end{equation} 
\indent where $\mathcal{L}_{\text{SMPL}}$ minimizes the absolute error on SMPL parameters $\hat{\Psi}$ and squared error on the 3D joint positions \textit{$\hat{S}$}. $\mathcal{L}_{\text{v2v}}$ (vertex-to-vertex loss) provides extra supervision to better match ground truth body mesh by minimizing squared error on the vertex positions \textit{$\hat{V}$}. $\mathcal{L}_{\text{P3D}}$ (3D pressure loss) minimizes the squared error on the 3D pressure map values.  $\mathcal{L}_{\text{contact}}$ is applied as a cross-entropy loss between predicted and ground truth 3D contact, where the ground truth 3D contact is obtained from all non-zero elements of the ground truth 3D pressure maps. The loss weighting coefficients are set empirically. Further details about the loss function are mentioned in appendix \cref{ssec:supp_bptrain}.

\subsection{BodyMAP-WS}

BodyMAP-WS is a variant of the BodyMAP model, learning without supervision for the 3D pressure map prediction. This model utilizes a pre-trained mesh regressor to obtain mesh predictions and ResNet image features as its primary inputs. The network architecture follows a similar design as BodyMAP’s pressure prediction part (\cref{fig:modelfull}) in its use of FIM and PointNet~\cite{qi2017pointnet}, and is further illustrated in the appendix \cref{fig:s_modelws}. To train BodyMAP-WS without supervision, we drop the prediction of binary contact value from BodyMAP and directly predict the pressure value at each vertex. During training, the network constructs a 2D projection of the predicted 3D pressure map and aligns it with the input pressure image, allowing for the implicit learning of the 3D applied pressure map onto the body mesh. To form the differentiable projection, we utilize the co-registration of the mesh with pressure taxels~\cite{clever2022bodypressure}, which establishes a direct link between the pressure on vertices $v_j$ and the pressure on the taxel located beneath the vertex. To project the predicted 3D pressure map onto a 2D image, we iterate over all the pressure taxels and compute the average pressure value of the vertices positioned directly above a taxel (vertices with $x$,~$y$ coordinates corresponding to the taxel). 

\subsubsection{Training strategy}

To form the pre-trained mesh regressor, we first train the mesh regressor section of BodyMAP using supervision with $\mathcal{L}_{\text{SMPL}}$ and $\mathcal{L}_{\text{v2v}}$. Subsequently, BodyMAP-WS is trained to leverage the frozen pre-trained mesh model's predicted ResNet features and mesh vertex locations as inputs, for predicting the 3D applied pressure map. This prediction is guided by the following loss function:

\begin{equation} \label{eq:2}
    L = \mathcal{L}_{\text{P2D}} + \lambda_{1}\mathcal{L}_{\text{Preg}}
\end{equation}

Here, $\mathcal{L}_{\text{P2D}}$ (2D pressure loss) minimizes the squared error between the 2D projection of the predicted 3D pressure map and the input pressure image. The bed mattress is situated on the $Z=0$ plane, and vertices predicted to be positioned above this plane ($Z > 0$) should ideally not have any applied pressure on them as they do not make contact with the mattress. To enforce this constraint, we utilize $\mathcal{L}_{\text{Preg}}$, a regularization term that penalizes positive pressure on these vertices by minimizing the norm of predicted pressure values for these vertices. Further details about the loss function are mentioned in appendix \cref{ssec:supp_bpws_train}.

\section{Evaluation}

The training datasets~\cite{clever2022bodypressure, liu2019seeing, liu2022simultaneously} used in this work provide aligned input depth and pressure images. Clever~\etal~\cite{clever2022bodypressure} provide 3D ground truth meshes for their released synthetic data and SMPL-3D fits for the real SLP dataset~\cite{liu2019seeing, liu2022simultaneously}.
To establish the ground truth 3D pressure map, we utilize the vertical projection method outlined in~\cite{clever2022bodypressure}. This method involves projecting the 2D pressure image onto the 3D-oriented ground truth mesh, forming a ground truth 3D pressure map. Although this method has limitations for self-contact (overlapping body
parts), this procedure provides effective approximations for
ground truth 3D pressure maps, facilitating comparison with prior methods~\cite{yin2022multimodal, clever2022bodypressure}.

We train our network, BodyMAP, on the entire simulated BodyPressureSD dataset~\cite{clever2022bodypressure} and the initial 80 subjects (1-80, with the exclusion of subject 7 due to calibration errors) from the real-world SLP dataset~\cite{liu2019seeing, liu2022simultaneously}. We pre-train the mesh regressor for BodyMAP-WS on the same training split. However, we train the BodyMAP-WS model on only the real-world SLP training data (1-80 subjects)~\cite{liu2019seeing, liu2022simultaneously} using only the depth and pressure images and no ground truth 3D pressure map data. We evaluate our methods on the final 22 subjects (81-102) from the real SLP dataset~\cite{liu2019seeing, liu2022simultaneously}, which aligns with the analysis of our baselines~\cite{clever2022bodypressure, yin2022multimodal}.

\subsection{Network Evaluation}

We evaluate our approaches using multiple 3D pose, 3D shape and 3D pressure map metrics. 

For 3D pose prediction, we evaluate our methods with 3D mean-per-joint position error (MPJPE) and 3D per-vertex error (PVE). These are computed as mean Euclidean error between predicted and ground truth 3D joint positions and 3D vertex positions respectively. 

For 3D shape prediction, we assess our methods by computing anatomical measurements for height and circumferences of the chest, waist, and hips of the mesh as proposed by Choutas~\etal~\cite{choutas2022accurate}. The measurements are computed on the SMPL~\cite{loper2023smpl} rest pose mesh, which is generated using predicted shape parameters and rest pose angle parameters. The 3D shape metrics are calculated as the absolute error between predicted and ground truth anatomical values.

We employ the below metrics to evaluate the 3D pressure map predictions:

\begin{itemize}
    \item Vertex-to-Vertex Pressure (v2vP) Error: This metric, as proposed by Clever~\etal~\cite{clever2022bodypressure}, assesses the alignment between the predicted 3D pressure map and the ground truth data at the vertex level.
    \item v2vP 1EA (v2vP one edge away) \& v2vP 2EA: The dense structure of the SMPL mesh~\cite{loper2023smpl} results in the 3D pressure map predicted at the vertex level to have a very fine granularity. To account for the region size over which pressure is applied, we form and evaluate coarse representations of the 3D pressure map. The coarse representation is obtained by setting the pressure value at each vertex as the average pressure value of its neighboring vertices. This averaging process can be interpreted as a smoothing operator on the vertex-level 3D pressure maps. For v2vP 1EA, we consider the vertices that are one edge away to calculate the averages. For v2vP 2EA, we include the vertices that are both one and two edges away. The smoothing operator region is illustrated in the appendix \cref{fig:s_ea}. After forming these coarser pressure maps for both the ground truth and predicted 3D pressure maps, we apply the v2vP metric to calculate the mean squared error between predicted and ground truth 3D pressure maps, resulting in v2vP 1EA and v2vP 2EA. 
\end{itemize}  
In addition, we assess pressure prediction error on a per-body part basis. To do so, we identify 14 distinct body parts, including left \& right heels, toes, elbows, shoulders, hips and spine, head, sacrum and ischium. We compute the v2vP metric on only the vertices belonging to a body part.

We compare our methods, BodyMAP and BodyMAP-WS, against the multimodal Pyramid Fusion model from Yin~\etal~\cite{yin2022multimodal} and BPB, BPW models from Clever~\etal~\cite{clever2022bodypressure}. We compute the above evaluation metrics for the BPW model, for which the model weights are made available by the authors. We report the metrics for the other methods when available in their respective papers. 

Furthermore, we conduct a custom training of models on only lateral poses (left \& right lateral), and evaluate on supine poses. We compare two BodyMAP-WS models, one trained on only lateral poses and the other trained on both lateral and supine poses, with BodyMAP and BPW~\cite{clever2022bodypressure} trained on only lateral poses. For BodyMAP-WS models, we freeze the mesh model once it is pre-trained on lateral poses. This evaluation allows us to test on an out-of-training-distribution setting and gauge the improvements from training BodyMAP-WS implicitly on the entire data.

Additionally, we compare our methods to a custom baseline, BodyMAP-Conv, that is similarly capable of jointly predicting human body mesh and a 3D applied pressure map. BodyMAP-Conv, illustrated in the appendix \cref{fig:s_modelC}, replaces the FIM and PointNet components of BodyMAP with fully connected layers to predict 3D pressure map $\hat{P}$.

\textbf{Hyperparameters}: We train our networks, BodyMAP and BodyMAP-Conv, with the loss function defined in \cref{eq:1} and $\lambda_{1}$ = 0.25,  $\lambda_{2}$ = 0.1, and $\lambda_{3}$ = 0.1. We train BodyMAP-WS with loss as defined in \cref{eq:2} and $\lambda_{1}$ = 500. We train all networks with a batch size of 64, learning rate of 0.0001, weight decay of 0.0005, and the Adam optimizer \cite{kingma2014adam} for gradient optimization and use random rotation and random erasing image augmentations. We train BodyMAP and BodyMAP-Conv for 100 epochs, while we train BodyMAP-WS for 15 epochs.

\begin{figure*}
\begin{center}
\includegraphics[width=1.0\linewidth]{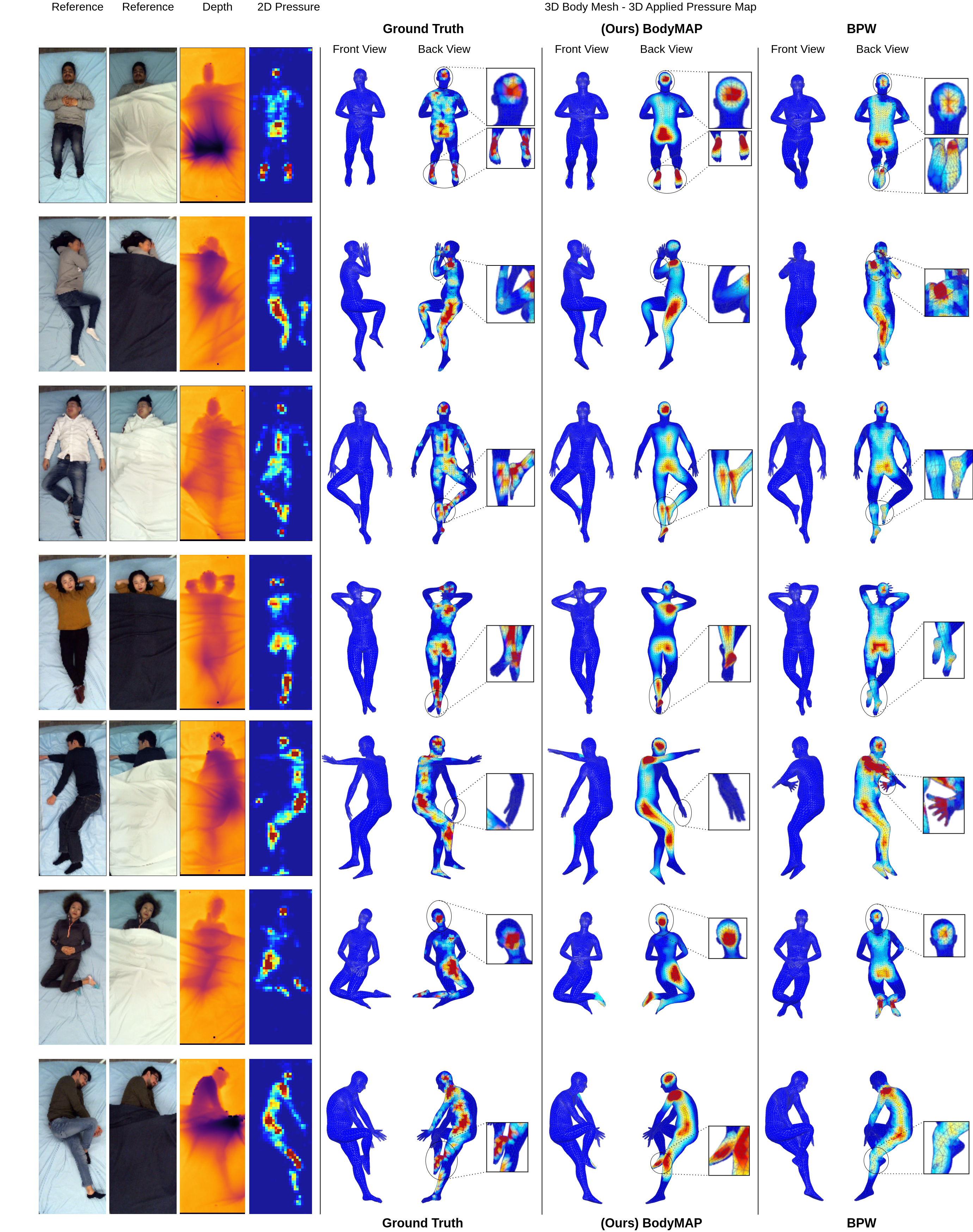}
\end{center}
   \caption{Results of inferring body mesh and 3D applied pressure map for examples from the SLP~\cite{liu2019seeing, liu2022simultaneously} test set.}
\label{fig:modelcomp}
\end{figure*}
\section{Results and Discussion}

\cref{tab:tab1_mainall} presents the results of BodyMAP compared to all baseline methods across various metrics for 3D pose, 3D shape, and 3D pressure distribution, evaluated on 22 test subjects from the real SLP dataset~\cite{liu2019seeing, liu2022simultaneously}. The averages are computed over all blanket configurations present in the SLP dataset~\cite{liu2019seeing, liu2022simultaneously}.

BodyMAP-PointNet, our model utilizing PointNet~\cite{qi2017pointnet} for 3D pressure map prediction, achieves notable improvements over the prior state-of-the-art BPW~\cite{clever2022bodypressure}. Specifically, when trained solely on the depth modality akin to the training approach of BPW~\cite{clever2022bodypressure}, BodyMAP-PointNet achieves a 12.5\% and 13\% enhancement for body mesh and 3D pressure map prediction tasks, respectively. When trained on multiple modalities, the BodyMAP-PointNet model attains a substantial 25\% improvement for both tasks over BPW~\cite{clever2022bodypressure}. Visual comparisons between BodyMAP-PointNet and BPW~\cite{clever2022bodypressure} are illustrated in \cref{fig:modelcomp}, and in the appendix in \cref{fig:s_c1}, and \cref{fig:s_c2}.

Remarkably, our BodyMAP-WS model demonstrates an 8.8\% performance improvement over BPW~\cite{clever2022bodypressure} for 3D pressure map prediction, despite lacking access to ground truth 3D pressure maps during training. Furthermore, our baseline model, BodyMAP-Conv, surpasses BPW~\cite{clever2022bodypressure} in direct comparison, trained solely on the depth modality. We hypothesize that errors introduced at each stage in BPW~\cite{clever2022bodypressure} accumulate, leading to adverse impacts on their final 3D pressure map predictions.

Detailed evaluations over specific blanket configurations are presented in the appendix \cref{tab:blanket_table}.

\begin{table*}[t!]

\begin{center}
\renewcommand{\arraystretch}{1.1}
\resizebox{\textwidth}{!}
{\begin{tabular} {|c|c||c|c||c|c|c|c||c|c|c|}

\hline
   \multicolumn{2}{|c||}{}&
   \multicolumn{2}{c||}{3D Pose Error (mm)~$\downarrow$}& 
   \multicolumn{4}{c||}{3D Shape Error (cm) $\downarrow$ }& 
   \multicolumn{3}{c|}{3D Pressure Error (kPa$^2$)~$\downarrow$ } \\

\hline
\hspace{-2mm} Network \hspace{-2mm}&   
{Modalities} &
{MPJPE}&
{PVE}&
{Height } &
{Chest } &
{Waist } &
{Hips } &
{v2vP}&
{v2vP 1EA}&
{v2vP 2EA} \\

\hline
\hline
Pyramid Fusion~\cite{yin2022multimodal} & RGB-D-PI-IR & 79.64 & - & - & - & - & - & \hspace{5mm}- \hspace{5mm} & - & -\\
\hline
BPBnet~\cite{clever2022bodypressure} & D & 74.54 & - & - & - & - & - &  2.87 & - & - \\
\hline
BPWnet~\cite{clever2022bodypressure} & D & 69.36 & 84.72 & 3.85 & 3.78 & 4.70 & 3.21 & 2.84 & 1.79 & 1.22 \\
\hline
BodyMAP - Conv & PI & 76.38~$\pm$~2.060  & 93.24~$\pm$~2.015 & 4.36~$\pm$~0.125 & \textbf{2.86~$\pm$~0.147} & 3.75~$\pm$~0.078 & 3.19~$\pm$~0.254 & 2.58~$\pm$~0.009 & 1.61~$\pm$~0.006 & 1.09~$\pm$~0.005 \\
\hline
BodyMAP - PointNet & PI & 76.54~$\pm$~3.170 & 93.6~$\pm$~4.029 & 4.83~$\pm$~0.164 & 3.06~$\pm$~0.323 & 4.1~$\pm$~0.433 & 3.56~$\pm$~0.409 & 2.29~$\pm$~0.014 & 1.39~$\pm$~0.011 & 0.92~$\pm$~0.007 \\
\hline
BodyMAP - Conv & D & 61.29~$\pm$~1.978 & 72.97~$\pm$~1.797 & 3.98~$\pm$~0.144 & 3.37~$\pm$~0.159 & 4.39~$\pm$~0.212 & 3.62~$\pm$~0.382 & 2.66~$\pm$~0.018 & 1.68~$\pm$~0.017 & 1.15~$\pm$~0.014 \\
\hline
BodyMAP - PointNet & D & 60.66~$\pm$~1.615 & 72.47~$\pm$~1.687 & 3.99~$\pm$~0.569 & 3.2~$\pm$~0.241 & 4.3~$\pm$~0.407 & 3.51~$\pm$~0.229 & 2.47~$\pm$~0.034 & 1.54~$\pm$~0.03 & 1.04~$\pm$~0.026 \\
\hline
BodyMAP - Conv & D~-~PI & 51.79~$\pm$~0.379 & 62.9~$\pm$~0.32 & 3.61~$\pm$~0.251 & 3.09~$\pm$~0.256 & 4.1~$\pm$~0.407 & 3.46~$\pm$~0.484 & 2.56~$\pm$~0.007 & 1.59~$\pm$~0.007 & 1.07~$\pm$~0.008 \\
\hline
BodyMAP - PointNet & D~-~PI & \textbf{51.01~$\pm$~1.071} & \textbf{61.66~$\pm$~0.919} & \textbf{3.49~$\pm$~0.459} & 2.93~$\pm$~0.215 & \textbf{3.7~$\pm$~0.071} & \textbf{2.8~$\pm$~0.360} & \textbf{2.14~$\pm$~0.030} & \textbf{1.27~$\pm$~0.025} & \textbf{0.83~$\pm$~0.020} \\
\hline

\end{tabular}}
\end{center}
\vspace{-18pt}
\caption{Results for 3D pose, 3D shape, and 3D pressure error metrics evaluated for the 22 test subjects in SLP dataset~\cite{liu2022simultaneously, liu2019seeing}, averaging over all blanket conditions (uncovered, cover1, and cover2). The modalities described are depth (D), pressure image (PI), infrared (IR) and RGB. The metrics for BodyMAP models are shown with the mean and standard deviation computed over 3 random runs.}
\label{tab:tab1_mainall}
\end{table*}


\begin{table*}[h]

\vspace{-1pt}
\begin{center}
\renewcommand{\arraystretch}{1.1}
\resizebox{\textwidth}{!}
{
\begin{tabular}{|c|c*{14}{|c}|c|}

\hline
   \multicolumn{2}{|c|}{}&
   \multicolumn{14}{c|}{3D Pressure Error v2vP (kPa$^2$)~$\downarrow$ }\\

\hline

\hspace{-2mm} Network \hspace{-2mm} &   
{Modalities}&
{Left Heel} &
{Right Heel} &
{Left Toes} &
{Right Toes} &
{Left Elbow} &
{Right Elbow} &
{Left Shoulder} &
{Right Shoulder} &
{Spine} &
{Head} &
{Left Hip} &
{Right Hip} &
{Sacrum}&
{Ischium}\\
\hline
\hline
BPW~\cite{clever2022bodypressure} & D & 4.13 & 3.75 & 0.41 & 0.36 & 1.55 & 1.41 & 1.83 & 1.65 & 1.76 & 13.06 & 15.83 & 18.29 & 10.60 & 6.11 \\
\hline
BodyMAP-Conv & D &	5.23$\pm$0.007 & 4.46$\pm$0.014 & \textbf{0.32$\pm$0.001} & 0.31$\pm$0.001 & 1.05$\pm$0.005 & 0.77$\pm$0.019 & 1.49$\pm$0.007 & 1.32$\pm$0.021 & 1.63$\pm$0.020 & 11.35$\pm$0.108 & 14.28$\pm$0.136 &	16.65$\pm$0.145 & 9.12$\pm$0.152 & 5.52$\pm$0.064 \\
\hline
BodyMAP-WS & D~-~PI &	4.26$\pm$0.164 & 3.74$\pm$0.009 & 0.42$\pm$0.022 & 0.36$\pm$0.02 & 1.46$\pm$0.015 & 1.1$\pm$0.063 & 1.71$\pm$0.075 & 1.6$\pm$0.037 & 1.67$\pm$0.017 & 11.4$\pm$0.361 & 13.13$\pm$0.251 &	17.72$\pm$0.919 & 7.6$\pm$0.252 & 5.15$\pm$0.057 \\
\hline
BodyMAP-Conv & D~-~PI &	5.22$\pm$0.015 & 4.44$\pm$0.009 & \textbf{0.32$\pm$0.001} & 0.32$\pm$0.001 & 1.05$\pm$0.021 & 0.77$\pm$0.007 & 1.41$\pm$0.047 & 1.25$\pm$0.043 & 1.54$\pm$0.026 & 11.22$\pm$0.199 & 13.25$\pm$0.07 &	15.76$\pm$0.026 & 8.6$\pm$0.16 & 5.24$\pm$0.116 \\
\hline
BodyMAP-PointNet & D~-~PI &	\textbf{3.81$\pm$0.088} & \textbf{3.46$\pm$0.016} & 0.33$\pm$0.005 & \textbf{0.31$\pm$0.002} & \textbf{0.82$\pm$0.009} & \textbf{0.61$\pm$0.019} & \textbf{1.25$\pm$0.053} & \textbf{1.16$\pm$0.019} & \textbf{1.52$\pm$0.007} & \textbf{10.22$\pm$0.269} & \textbf{10.55$\pm$0.233} &	\textbf{13.44$\pm$0.205} & \textbf{7.17$\pm$0.187} & \textbf{4.49$\pm$0.180} \\	
\hline
\end{tabular}
}
\end{center}
\vspace{-18pt}
\caption{Results of 3D pressure map error evaluated over multiple body regions for the 22 test subjects in SLP dataset~\cite{liu2019seeing, liu2022simultaneously}, averaging over all blanket conditions (uncovered, cover1, and cover2). Our method achieves more than 30\% improvement for body regions vulnerable to pressure injuries such as left hip \& sacrum regions. The metrics are shown with the mean and std. dev. computed over 3 random runs.}
\label{tab:tab2_detailedpressure}
\end{table*}

\textbf{Elevated precision in 3D pressure map prediction across critical body regions}: Specific anatomical regions, such as the back of the head, heels, sacrum, spine, ischium, and hips, are known to be particularly susceptible to pressure ulcers~\cite{8755997}. Evaluating model performance across multiple body regions, revealed substantial improvements over BPW~\cite{clever2022bodypressure}. The results are summarized in \cref{tab:tab2_detailedpressure}.

BodyMAP-WS utilizes a regularization loss that discourages the model from predicting pressure on body vertices not in contact with the mattress (\cref{eq:2}). As a possible consequence, the model exhibits slightly lower performance than BPW~\cite{clever2022bodypressure} in areas with smaller contact surfaces, notably the heels and toes. However, in regions like the hips and spine, BodyMAP-WS surpasses BPW~\cite{clever2022bodypressure}.

BodyMAP-PointNet achieves the lowest pressure error across all body parts. These advancements bear significant implications for aiding caregivers in localizing peak pressure on critical body regions, potentially enhancing the effectiveness of pressure injury prevention protocols. 

\textbf{Implicit learning of 3D applied pressure map eliminates the need for ground truth annotations}: BodyMAP-WS, is trained through alignment of the 2D projection of the predicted 3D pressure map with the input pressure image. As outlined in \cref{tab:ws_side_table}, when trained on lateral poses, and evaluated on supine poses, BodyMAP-WS outperforms the prior state-of-the-art BPW method~\cite{clever2022bodypressure}. Moreover, as presented in \cref{tab:ws_side_table}, extending the training of BodyMAP-WS to the entire dataset (comprising both supine and lateral poses), while continuing to use the mesh model pre-trained on only lateral poses, leads to further performance gains, even surpassing our proposed BodyMAP-PointNet model. 
This methodology to train models without relying on ground truth labels has the potential to streamline cost-intensive annotation efforts, potentially unlocking training these models on unlabeled data from deployed sensors.

\textbf{Multimodal input elevates performance}: The results, as presented in ~\cref{tab:tab1_mainall} for BodyMAP-PointNet and BodyMAP-Conv with various modality configurations, underscore the significance of combining pressure and depth modalities for optimal results. Notably, there is a significant reduction in body mesh prediction performance when exclusively using the pressure modality. This decline is attributed to the absence of visual cues from pressure images when limbs are not in contact with the mattress. Conversely, we observe the pressure modality to be more effective for 3D pressure map prediction when compared to the depth modality. This substantiates a preference to leverage both the depth (top-down view) and 2D pressure mattress images (bottom-up view), providing greater context for the model to jointly predict body mesh and 3D applied pressure maps.

Moreover, as illustrated in the appendix \cref{tab:blanket_table}, integrating multiple modalities leads to consistent mesh prediction performance across all blanket configurations. This stands in contrast to BPW~\cite{clever2022bodypressure}, which experiences a significant drop when individuals in bed are covered with blankets due to its training on only a depth modality.

\textbf{Feature Indexing Module improves 3D pressure performance}: The proposed Feature Indexing Module (FIM) creates a comprehensive feature set for each vertex by fusing vertex locations, ResNet features (formed before the global average pool), and input image features. This provides the model with rich contextual information, enhancing the 3D applied pressure map prediction. Notably, fusing global ResNet features (formed after the global average pool) with PointNet encoder features also yields an improvement in predicting 3D pressure maps. We present the ablations of FIM in the appendix \cref{tab:tab4_fim}. 

\begin{table}[t]

\begin{center}

\renewcommand{\arraystretch}{1.1}

\resizebox{\columnwidth}{!}{\begin{tabular}{|c|c|c|c|}

\hline
{Network} &
{Train Data} &
{3D Pose Error} &
{3D Pressure Error} \\
& &
{MPJPE (mm) $\downarrow$} &
{v2vP (kPa$^2$)~$\downarrow$} \\
\hline
\hline
BPW~\cite{clever2022bodypressure} & Lateral poses & 122.8 & 3.164 \\
\hline
BodyMAP-PointNet & Lateral poses & 100.96 & 2.614 \\
\hline
BodyMAP-WS & Lateral (mesh) + Lateral (pressure) & \textbf{98.82} & 2.865 \\
\hline
BodyMAP-WS & Lateral (mesh) + All (pressure) & \textbf{98.82} & \textbf{2.513} \\
\hline
\end{tabular}}
\end{center}
\vspace{-18pt}
\caption{Results from training on lateral poses and evaluating on supine poses in the test subjects in the SLP dataset~\cite{liu2019seeing, liu2022simultaneously}, averaging over all blanket conditions (uncovered, cover1 and cover2). Here, BodyMAP-WS models use a mesh model pre-trained solely on lateral poses. All models are trained on only depth modality.}
\label{tab:ws_side_table}
\end{table}

\section{Conclusion}

Our proposed method, BodyMAP, jointly predicts the body mesh and 3D applied pressure map for people in bed. These predictions offer interpretable visualizations of pressure on the body, which could aid caregivers in accurately identifying high-pressure regions and potentially enhance the prevention of pressure injuries. Furthermore, we introduced BodyMAP-WS to implicitly learn a 3D pressure map on the body without relying on labeled ground truth data, opening new avenues for learning on large real-world datasets where obtaining 3D labels is both costly and challenging. We have demonstrated the performance of our methods in a series of ablations and comparisons to prior state-of-the-art models.

{
    \small
    \bibliographystyle{ieeenat_fullname}
    \bibliography{main}
}

\clearpage
\maketitlesupplementary

\section{Input Modalities}
\label{sec:supp_modalities}

\begin{figure}[t]
\begin{center}
\includegraphics[width=1\linewidth]{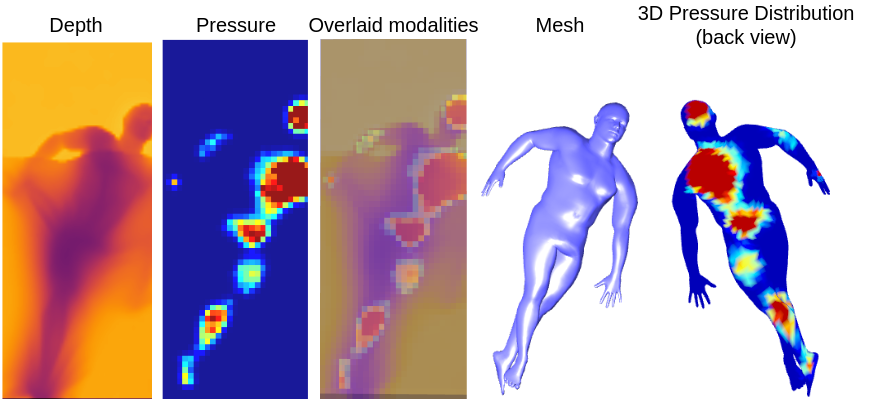}
\end{center}
   \caption{Depth and pressure image complement each other. Image of overlaid modalities (third column) depicts the enhanced context available to the model.}
\label{fig:smod}
\end{figure}

BodyMAP uses as inputs the depth image and the corresponding pressure image for an individual in bed. It jointly predicts both the body mesh (3D pose \& shape) and the pressure applied on the body, in the form of a 3D pressure map. As illustrated in \cref{fig:smod}, the depth image provides a top-to-bottom view, and the pressure image provides a `bottom-to-up' view, capturing complementary features of the body. Using these two visual modalities enhances the model's context, allowing it to accurately predict the body mesh and 3D pressure map, even when the individual is heavily covered with blankets. 

\section{BodyMAP}
\label{sec:supp_bodymapmodel}

\subsection{BodyMAP-PointNet}
\label{ssec:supp_bppoint}

BodyMAP-PointNet substantially surpasses prior methods by 25\% on both body mesh and 3D pressure map prediction tasks, for in-bed, visually occluded people. Illustrated in \cref{fig:modelfull}, the model uses uses a ResNet18~\cite{he2016deep} and MLP to predict the SMPL~\cite{loper2023smpl} parameters \textit{$\hat{\Psi}$}. These parameters include body shape, joint angles, root-joint translation, and root-joint rotation: $\hat{\Psi} = [\hat{\beta}, \hat{\Theta}, \hat{s}, \hat{x}, \hat{y}]$. We use the SMPL embedding block~\cite{kanazawa2018end} to obtain the SMPL~\cite{loper2023smpl} body mesh. The model then leverages the proposed Feature Indexing Module to form features for each vertex of the predicted mesh. Subsequently, it uses a PointNet~\cite{qi2017pointnet} model to infer the 3D pressure map. This new method unifies the design for jointly predicting body mesh and 3D pressure map. The feature extractors used in our model, ResNet and PointNet, are easily replaceable with other modern feature extractors. 

\subsection{BodyMAP-Conv}
\label{ssec:supp_bpconv}

\begin{figure}[t]
\begin{center}
\includegraphics[width=1.0\linewidth]{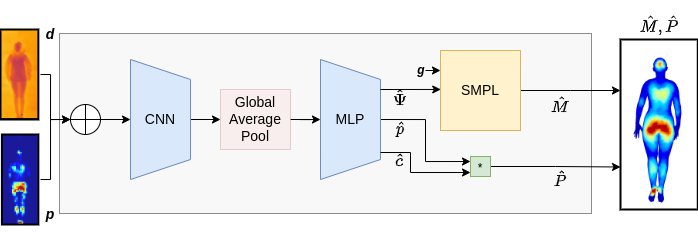}
\end{center}
   \caption{BodyMAP-Conv jointly predicts body mesh and 3D applied pressure map for an individual in bed. The model replaces the FIM and PointNet components of BodyMAP-PointNet, and instead predicts the 3D pressure map using an MLP.} 
\label{fig:s_modelC}
\end{figure}

We develop a baseline method denoted as BodyMAP-Conv, depicted in \cref{fig:s_modelC}. BodyMAP-Conv replaces FIM and PointNet with fully connected layers to predict the 3D pressure map jointly with the SMPL parameters. This model design allows us to evaluate the enhancements provided by FIM and PointNet for 3D pressure map prediction. 

\subsection{Training Strategy}
\label{ssec:supp_bptrain}

We train BodyMAP-PointNet and BodyMAP-Conv to jointly predict the body mesh and 3D pressure map with the following loss function: 
\begin{equation} 
    L = \mathcal{L}_{\text{SMPL}} + \lambda_{1}\mathcal{L}_{\text{v2v}} + \lambda_{2}\mathcal{L}_{\text{P3D}} + \lambda_{3}\mathcal{L}_{\text{contact}}
\label{eq:s3}
\end{equation}

Here, $\mathcal{L}_{\text{SMPL}}$ minimizes the absolute error on SMPL parameters $\hat{\Psi}$ and squared error on the 3D joint positions.

\begin{equation} 
\begin{split}
    \mathcal{L}_{\text{SMPL}} & = \frac{1}{N_\beta \sigma_\beta}\lVert \hat{\beta} - \beta \rVert_{1} +  
    \newline \frac{1}{N_\Theta \sigma_\Theta}\lVert \hat{\Theta} - \Theta \rVert_{1} + \\ & 
    \frac{1}{6\sigma_yx}( \lVert \hat{x} - x \rVert_{1}  + \lVert \hat{y} - y \rVert_{1}) + 
    \frac{1}{N_s \sigma_s}\sum_{j=1}^{N_s} \lVert \hat{s}_j - s_j \rVert_{2}
\label{eq:s4}
\end{split}
\end{equation}

where \textit{$N_{\beta}$} = 10 body shape parameters, \textit{$N_{\Theta}$} = 69 joint angle parameters, \textit{$x$} and \textit{$y$} represent global rotation, and \textit{$N_{s}$} = 24 joint positions in 3D with \textit{$s_j$} representing each joint position. Each term is normalized by their respective standard deviations computed over the entire synthetic dataset.~\cite{clever2022bodypressure}.  

$\mathcal{L}_{\text{v2v}}$ (vertex-to-vertex loss) minimizes the Euclidean error between the predicted and ground truth vertex positions, normalized by the standard deviation \textit{$\sigma_{v}$} calculated over the entire synthetic data, and is calculated as: 

\begin{equation}
\mathcal{L}_{\text{v2v}} = \frac{1}{N_{v}\sigma_{v}}\sum_{j=1}^{N_{v}}|| \hat{v}_{j} - v_{j} ||_{2}
\label{eq:s5}
\end{equation}

Here $N_v$ represents the number of vertices for the body mesh which is equal to 6890 for SMPL~\cite{loper2023smpl} meshes.

$\mathcal{L}_{\text{P3D}}$ (3D pressure map loss) minimizes the squared error between predicted and ground truth per-vertex pressure value, normalized by the standard deviation of ground truth pressure map \textit{$\sigma_{p}$} calculated over the entire synthetic data, and is modeled as:

\begin{equation}
\mathcal{L}_{\text{P3D}} = \frac{1}{N_{v}\sigma_{p}}\sum_{j=1}^{N_{v}}|| \hat{p}_{j} - p_{j} ||_{2}
\label{eq:s6}
\end{equation}

$\mathcal{L}_{\text{contact}}$ is calculated as a cross-entropy loss between the predicted and ground truth per-vertex binary contact values, where the ground truth contact is obtained from all non-zero elements of ground truth 3D pressure maps.  
\begin{equation}
\mathcal{L}_{\text{contact}} = \frac{1}{N_{v}\sigma_{c}}\sum_{j=1}^{N_{v}}-{(y_{j}\log(p_{j}) + (1 - y_{j})\log(1 - p_{j}))}
\label{eq:s7}
\end{equation}
Here \textit{$y_{j}$} represents the ground truth contact and \textit{$p_j$} represents the predicted probability of contact value, calculated for each vertex. \textit{$\sigma_{c}$} represents the standard deviation of ground truth contact calculated over the entire synthetic data~\cite{clever2022bodypressure}.

\textbf{Hyperparameters}: We empirically set $\lambda_{1}$ = 0.25,  $\lambda_{2}$ = 0.1, and $\lambda_{3}$ = 0.1 to train the models, BodyMAP-PointNet and BodyMAP-Conv, with the loss function defined in \cref{eq:s3}

\begin{table*}[t!]

\vspace{-5mm}
\begin{center}
\renewcommand{\arraystretch}{1.1}

\resizebox{\textwidth}{!}{\begin{tabular} {|c|c||c|c|c|c||c|c|c|c||c|c|c}
\hline
   \multicolumn{2}{|c||}{}&
   \multicolumn{4}{c||}{3D Pose Error - MPJPE (mm) $\downarrow$ }& 
   \multicolumn{4}{c||}{3D Pressure Distribution Error - v2vP (kPa$^2$) $\downarrow$ } \\

\hline
 \hspace{-2mm} Network \hspace{-2mm}&   
{Modalities}&
{Uncovered  }
{}&
{Cover 1}
{}&
{Cover 2  }
{ }& 
{Overall }
{}&
{Uncovered }
{}&
{Cover 1 }
{}&
{Cover 2}
{ }&
{Overall }
{ }\\

\hline
\hline
Pyramid Fusion~\cite{yin2022multimodal} \hspace{-1mm} & RGB-D-PI-IR\hspace{-1mm} & \hspace{-2mm} 78.80\hspace{-1mm} & \hspace{-2mm} 79.92\hspace{-1mm} & \hspace{-2mm} 80.21\hspace{-1mm} & \hspace{-2mm} 79.64\hspace{-1mm} & - & - & - & - \\ \hline
BPBnet~\cite{clever2022bodypressure} & D & 70.16 & 76.99 & 76.49 & 74.54 & 2.87 & 2.88 & 2.87 & 2.87 \\ \hline
BPWnet~\cite{clever2022bodypressure} & D & 63.64 & 72.4 & 72.04 & 69.36 & 2.85 & 2.85 & 2.82 & 2.84 \\ \hline									
BodyMAP - Conv & D & 55.7~$\pm$~1.753 & 64.34~$\pm$~2.099 & 63.82~$\pm$~2.117 & 61.29~$\pm$~1.978 & 2.68~$\pm$~0.019 & 2.66~$\pm$~0.02 & 2.63~$\pm$~0.017 & 2.66~$\pm$~0.018 \\ \hline
BodyMAP - Conv & D~-~PI & 49.48~$\pm$~0.127 & 52.9~$\pm$~0.576 & 53.0~$\pm$~0.521 & 51.79~$\pm$~0.379 & 2.59~$\pm$~0.008 & 2.55~$\pm$~0.007 & 2.53~$\pm$~0.007 & 2.56~$\pm$~0.007 \\ \hline
BodyMAP - PointNet & D~-~PI & \textbf{48.77~$\pm$~0.981} & \textbf{51.88~$\pm$~1.142} & \textbf{52.36~$\pm$~1.092} & \textbf{51.01~$\pm$~1.071} & \textbf{2.16~$\pm$~0.035} & \textbf{2.13~$\pm$~0.029} & \textbf{2.12~$\pm$~0.026} & \textbf{2.14~$\pm$~0.030} \\ \hline

\end{tabular}}
\end{center}
\vspace{-18pt}

\caption{Results for 3D pose error (MPJPE) \& 3D pressure map error (v2vP) for the 22 test subjects from SLP dataset, over multiple blanket conditions - uncovered (no blanket), cover1 (light blanket), cover2 (heavy blanket) and overall (average case). The modalities used are: depth image (D), pressure image (PI), infrared images (IR) and RGB images. The metrics for BodyMAP models are shown with the mean and standard deviation computed over 3 random runs.}
\label{tab:blanket_table} 
\end{table*}

\begin{figure*}
\begin{center}
\includegraphics[width=1.0\linewidth]{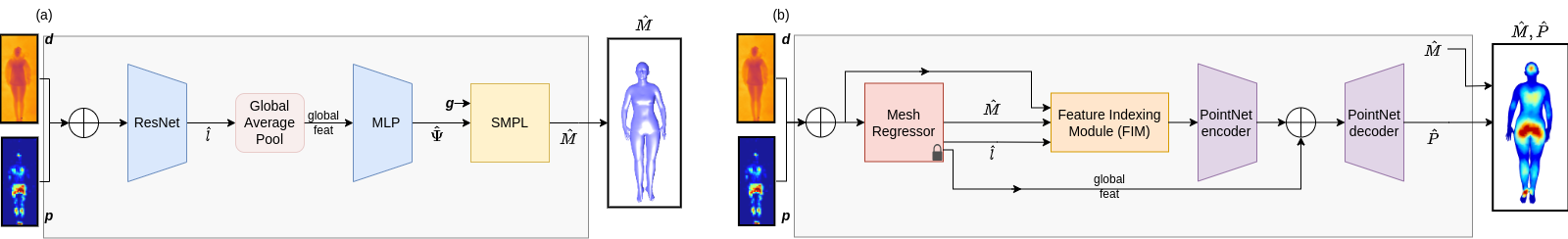}
\end{center}
   \caption{BodyMAP-WS implicitly learns to predict 3D pressure maps without using direct supervision. (a) The mesh regressor model is used as a frozen pre-trained model for BodyMAP-WS. (b) BodyMAP-WS leverages FIM and PointNet to directly predict the 3D pressure map (without predicting per-vertex binary contact like BodyMAP-PointNet).}
\label{fig:s_modelws}
\end{figure*}

\section{BodyMAP-WS}
\label{sec:supp_bpws}

We introduce BodyMAP-WS, designed to implicitly learn 3D pressure maps without using direct supervision. BodyMAP-WS leverages a pre-trained body mesh regressor network and trains to predict a 3D pressure map. The mesh regressor network used in our work is depicted in \cref{fig:s_modelws}(a). Using the mesh estimates, BodyMAP-WS leverages FIM and PointNet to predict the 3D pressure map. Unlike BodyMAP-PointNet which predicts the 3D pressure map as a product of the per-vertex pressure and binary contact value, BodyMAP-WS, depicted in \cref{fig:s_modelws}(b), directly predicts the 3D pressure map modeled by the per-vertex pressure value. 

For training, we form a differentiable 2D projection of the predicted 3D pressure map and align it with the input pressure image by minimizing loss between the formed projection and the input pressure image. This alignment allows the model to learn the correct 3D pressure map without any supervision with ground truth data. This training methodology allows the model to learn from unlabelled data, where collecting labels can be challenging and costly. 

\subsection{Training strategy}
\label{ssec:supp_bpws_train}

We pre-train the mesh regressor for BodyMAP-WS with the following loss function: 
\begin{equation} 
    L = \mathcal{L}_{\text{SMPL}} + \lambda_{1}\mathcal{L}_{\text{v2v}}
\label{eq:s9}
\end{equation}

The mesh regressor model is then frozen and is used to train BodyMAP-WS with the loss function:
\begin{equation}
    L = \mathcal{L}_{\text{P2D}} + \lambda_{1}\mathcal{L}_{\text{Preg}}
\label{eq:s10}
\end{equation}

Here, $\mathcal{L}_{\text{P2D}}$ (2D pressure loss) minimizes the squared error between the 2D projection of the predicted 3D pressure map and the input pressure image. 
\begin{equation} 
    \mathcal{L}_{\text{P2D}} = \lVert \hat{p} - p \rVert_{2}
\label{eq:s11}
\end{equation}

where \textit{$p$} is the input pressure image and \textit{$\hat{p}$} is the 2D projection of the predicted 3D pressure map. 

We model $\mathcal{L}_{\text{Preg}}$ (pressure regularization loss) as a regularization loss to penalize pressure predicted on vertices that lie above the mattress plane, as these are not in contact with the mattress. 
\begin{equation} 
    \mathcal{L}_{\text{Preg}} = \lVert \sum_{j=1}^{N_{\text{nocv}}}\hat{p_j} \rVert_{2}
\label{eq:s12}
\end{equation}

where \textit{$p_j$} represents pressure on a vertex not in contact with the mattress, \textit{$N_{\text{nocv}}$} represents the number of no-contact-vertices, i.e. vertices lying above the mattress plane ($Z = 0$) and is obtained as:
\begin{equation} 
    V_{\text{nocv}} = \{ V | V_{z} > 0 \}
\label{eq:s13}
\end{equation}
\begin{equation} 
    N_{\text{nocv}} = \left|V_{\text{nocv}}\right|
\label{eq:s13b}
\end{equation}


\textbf{Hyperparameters}: For pre-training the mesh regressor we set $\lambda_{1}$ = 0.25 in \cref{eq:s9}. To train BodyMAP-WS, we set $\lambda_{1}$ = 500 in \cref{eq:s10}. These hyperparameter values are set empirically. 

\begin{figure}[t]
\begin{center}
\includegraphics[width=1\linewidth]{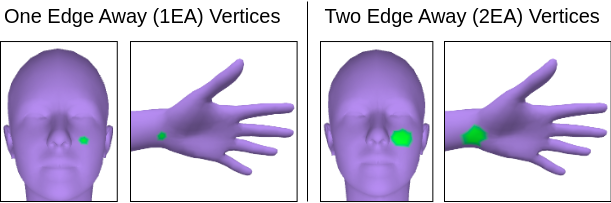}
\end{center}
   \caption{Smoothing region (shown in green color) of two vertices for one edge away (1EA) and two edge away (2EA) scenarios.}
\label{fig:s_ea}
\end{figure}

\section{Pressure Metrics Details}
\label{sec:supp_ea}

To evaluate predicted 3D pressure maps, we have introduced two metrics: v2vP 1EA (one edge away) and v2vP 2EA. These metrics calculate the squared error between the predicted and ground truth 3D pressure at coarser levels. These coarser 3D pressure maps are formed by setting the pressure at a vertex as the average of its neighboring vertices. We illustrate the spread of neighboring vertices used for v2vP 1EA and v2vP 2EA in \cref{fig:s_ea}.

\section{Further Results}
\label{sec:supp_res}

BodyMAP surpasses prior methods in both body mesh and 3D pressure map prediction. \cref{tab:blanket_table} presents detailed results over different blanket thickness configurations, evaluated over the 22 test subjects from the SLP dataset~\cite{liu2019seeing, liu2022simultaneously}. 

While BPW~\cite{clever2022bodypressure}, the previous state-of-the-art method, exhibits strong performance in body mesh prediction in scenarios where the individual is uncovered (uncovered case), its effectiveness notably diminishes when occlusions from blankets (cover1 and cover2) are introduced. In contrast, our methods, BodyMAP-PointNet and BodyMAP-Conv, when trained on multiple modalities, maintain consistent performance across all three blanket configurations, showcasing a significant improvement over BPW~\cite{clever2022bodypressure}. This evaluation reinforces the intuition that depth images capture limited details about the human body when individuals are covered with blankets. Instead, our approach of leveraging both modalities, with depth images providing the top-down view and 2D pressure images offering a `bottom-to-up' view, yields robust performance across different blanket configurations.

\begin{table}[t!]


\renewcommand{\arraystretch}{1.5}
\resizebox{\columnwidth}{!}
{\begin{tabular} {|c|c|c|c|c|c|c|c|c|c|}

\hline

   \multicolumn{4}{|c|}{BodyMAP - Feature Indexing Module Ablation}& 

   \multicolumn{3}{c|}{3D Pressure Error (kPa$^2$)~$\downarrow$} \\
\hline

{Vertex Location}&
{\hspace{2mm}Input Image \hspace{2mm}} &
{\hspace{2mm}ResNet Features \hspace{2mm}} &
{\hspace{-1mm}Global ResNet features\hspace{-1mm}} &
{\hspace{5mm}v2vP\hspace{5mm} } &
{v2vP 1EA} &
{v2vP 2EA} \\

\hline
\hline
	-&	\checkmark & -&	-&	2.765&	1.797&	1.256 \\
 \checkmark & - & - & - & 2.478 & 1.552 & 1.043 \\
 - & \checkmark &	- & \checkmark &	2.455 &	1.526 &	1.033 	\\																
 -&	\checkmark & \checkmark &	- &	2.369 &	1.419 &	0.921 \\	
 \checkmark & \checkmark & - & - & 2.334 & 1.442 & 0.969 \\
 \checkmark & - & - & \checkmark & 2.339 & 1.429 & 0.945 \\
 \checkmark & - & \checkmark & - & 2.205 & 1.311 & 0.855 \\
 \checkmark & - & \checkmark & \checkmark & 2.204 & 1.311 & 0.857 \\
 \checkmark & \checkmark & - & \checkmark & 2.195 & 1.315 & 0.862 \\
 \checkmark & \checkmark & \checkmark & - & 2.164 & 1.280 & 0.831 \\
 \checkmark & \checkmark & \checkmark & \checkmark & \textbf{2.130} & \textbf{1.260} & \textbf{0.822} \\
\hline

\end{tabular}}
\vspace{-0.1cm}

\caption{Ablation study of Feature Indexing Module (FIM) and use of global ResNet features (formed after global average pooling) for 3D pressure map prediction. These results are evaluated for BodyMAP-PointNet model trained with depth and pressure images as input modalities, averaging over all blanket configurations.}
\label{tab:tab4_fim}
\vspace{-0.6cm}
\end{table}

\textbf{Feature Indexing Module improves 3D pressure performance}: \cref{tab:tab4_fim} showcases the impact of different approaches to forming mesh features through the proposed Feature Indexing Module (FIM) and the use of global ResNet features (formed after global average pooling). The proposed fusing of vertex locations, accumulated input image and ResNet features (before global average pooling) through FIM, and global ResNet features, provides a comprehensive feature set for the model to learn from and result in accurate 3D pressure map predictions.

\textbf{Ablating training components reduces performance}: Ablating the $\mathcal{L}_{\text{v2v}}$ and $\mathcal{L}_{\text{contact}}$ loss functions from \cref{eq:s3} in BodyMAP training results in a performance drop in both body mesh and 3D pressure map prediction. \cref{tab:loss_table} showcases the outcomes across the test set, where we observe a performance improvement from predicting 3D contact and minimizing loss for contact and vertex locations. 

We also ablated training-time data augmentations and observed improvement in results from using random rotation and random erasing augmentations. \cref{tab:aug_table} outlines these results.

\section{Computational Analysis}
\label{sec:comp_analysis}

We conducted a comparative analysis between BodyMAP-PointNet, BodyMAP-Conv, and the prior state-of-the-art method BPW~\cite{clever2022bodypressure}, focusing on computational efficiency metrics including FLOP (total floating-point operations), parameter count, and inference time. The summarized results are detailed in \cref{tab:compute_table}.

\textbf{Parameter count}: BPW~\cite{clever2022bodypressure} employs two ResNet34~\cite{he2016deep} models to predict SMPL~\cite{loper2023smpl} parameters for body mesh formation, alongside a white-box model and a compact convolutional network to estimate the 2D pressure image. This multi-model approach results in BPW~\cite{clever2022bodypressure} containing a total of 42.73 million parameters. In contrast, our models, BodyMAP-Conv and BodyMAP-PointNet, adopt simpler architectures with a reduced parameter count compared to BPW~\cite{clever2022bodypressure}.

BodyMAP-Conv utilizes a ResNet18~\cite{he2016deep} and fully connected layers to predict jointly the SMPL~\cite{loper2023smpl} mesh parameters and a pressure value for each vertex, leading to an output size of 6890 for the entire SMPL~\cite{loper2023smpl} mesh, resulting in a total parameter count of 34.03 million.

BodyMAP-PointNet also employs a ResNet18~\cite{he2016deep} and fully connected layers for body mesh parameters and per-vertex pressure value prediction but optimizes for efficiency by sharing weights of the fully connected layers across vertices using the PointNet~\cite{qi2017pointnet} architecture. This design choice results in a highly efficient model with a significantly reduced parameter count of only 13.56 million.

\textbf{FLOPs}: Although BPW~\cite{clever2022bodypressure} uses three separate neural-network models, it only a predicts 2D pressure image and not a per-vertex 3D pressure map through the models. This design choice allows BPW~\cite{clever2022bodypressure} to use a minimal number of fully connected layers, resulting in the lowest total of 1.25 giga FLOPs. In contrast, both BodyMAP-Conv and BodyMAP-PointNet predict a 3D pressure map, employing fully connected layers, resulting in higher total FLOPs. 

BodyMAP-Conv utilizes a fully connected layer to predict a pressure value for each of the 6890 body mesh vertices, resulting in a total of 1.8 giga FLOPs. 

BodyMAP-PointNet leverages a PointNet~\cite{qi2017pointnet} model to predict the 3D pressure map. This requires the model to compute mesh features (features for each of the 6890 body mesh vertices) in many layers of the model, resulting in a total of 2.26 giga FLOPs. Despite the higher computational demand during inference, BodyMAP-PointNet's streamlined design results in lower memory requirements, facilitating easier scalability for jointly predicting body mesh and 3D pressure map.

\textbf{Inference time}: BPW~\cite{clever2022bodypressure} approximates the 3D pressure map by projecting pressure from the predicted 2D pressure image to the vertices of the predicted body mesh, necessitating vertex-wise iterations and resulting in an inference time of 3.5168 seconds. In contrast, BodyMAP-Conv and BodyMAP-PointNet predict the 3D pressure map directly, leading to significantly reduced inference times. 

BodyMAP-Conv utilizes a fully connected layer to predict the 3D pressure map, resulting in the lowest inference time of 0.0042 seconds.

Employing the PointNet~\cite{qi2017pointnet} architecture, BodyMAP-PointNet directly predicts the 3D pressure map in only 0.0046 seconds, demonstrating comparable inference time to BodyMAP-Conv.

\textbf{Analysis}: BodyMAP-PointNet demonstrates a substantial 68\% reduction in the number of parameters and a 99\% decrease in inference time compared to the prior state-of-the-art BPW~\cite{clever2022bodypressure}. These computational advantages translate to simplified scalability and real-time performance, crucial for implementation and deployment in hospitals and nursing facilities.

The joint prediction of body mesh and 3D pressure map for a person in bed by BodyMAP could empower caregivers with visualization of pressure distribution on the body. The low inference time required by BodyMAP allows for generating these visualizations in real-time. This eliminates the need for caregivers to manually assess pressure information, allowing them to prioritize providing care. These benefits could potentially enhance the pressure injury prevention process.

\begin{table}[t]

\vspace{-5mm}
\begin{center}

\renewcommand{\arraystretch}{1.1}

\resizebox{\columnwidth}{!}{\begin{tabular}{|c|c|c|}

\hline
   \multicolumn{1}{|c|}{BodyMAP-PointNet loss ablation}&
   \multicolumn{1}{c|}{3D Pose Error}& 
   \multicolumn{1}{c|}{3D Pressure Map Error} \\
\hline

{Extra Losses - v2v, contact} &
{MPJPE (mm) $\downarrow$} &
{v2vP (kPa$^2$) $\downarrow$} \\
\hline
\hline
- & 55.34 & 2.182 \\
\checkmark & \textbf{52.63} & \textbf{2.130} \\
\hline
\end{tabular}}
\end{center}
\vspace{-0.5cm}
\caption{Ablation study of extra losses, $\mathcal{L}_{\text{v2v}}$ (\cref{eq:s5}) and $\mathcal{L}_{\text{contact}}$ (\cref{eq:s7}),  used for training BodyMAP-PointNet. Here the model is trained with depth and pressure image as input modalities, with evaluation conducted on the test set from the SLP dataset~\cite{liu2019seeing, liu2022simultaneously}, averaging over all blanket configuration (uncovered, cover1 \& cover2)}
\label{tab:loss_table}
\end{table}

\begin{table}[t]

\vspace{-2mm}
\begin{center}

\renewcommand{\arraystretch}{1.1}

\resizebox{\columnwidth}{!}{\begin{tabular}{|c|c|c|c|}

\hline
   \multicolumn{2}{|c|}{BodyMAP-PointNet Augmentation ablation}&
   \multicolumn{1}{c|}{ 3D Pose Error}& 
   \multicolumn{1}{c|}{3D Pressure Map Error} \\
\hline

{Modality} &
{Data Augmentation} &
{MPJPE (mm) $\downarrow$ } &
{v2vP (kPa$^2$) $\downarrow$  } \\
\hline
\hline
D & - & 69.73 & 2.609 \\
D & \checkmark & 59.44 & 2.473 \\
D - PI & - & 60.07 & 2.277 \\
D - PI & \checkmark & \textbf{52.63} & \textbf{2.130}\\
\hline
\end{tabular}}
\end{center}
\vspace{-0.5cm}
\caption{Ablation study of data augmentations used for training BodyMAP-PointNet. The evaluation is conducted on the test set from the SLP dataset~\cite{liu2019seeing, liu2022simultaneously}, averaging over all blanket configurations (uncovered, cover1 \& cover2)}
\label{tab:aug_table}
\end{table}

\begin{table}[t]

\vspace{-2mm}
\begin{center}

\renewcommand{\arraystretch}{1.1}

\resizebox{\columnwidth}{!}{\begin{tabular}{|c|c|c|c|}

\hline

{Network} &
{FLOPs (G) $\downarrow$} &
{Params (M) $\downarrow$} &
{Inference Time (s) $\downarrow$} \\
\hline
\hline
BPW~\cite{clever2022bodypressure} & \textbf{1.2469} & 42.7316 & 3.516813 \\
BodyMAP-Conv & 1.8044 & 34.0252 & \textbf{0.004224} \\
BodyMAP-PointNet & 2.2619 & \textbf{13.5580} & 0.004650 \\
\hline
\end{tabular}}
\end{center}
\vspace{-0.5cm}
\caption{Results from model computation comparison. The evaluation is computed on a machine using Intel Core i9-13900KF-3 GHz-24 core CPU processor and NVIDIA GeForce RTX 4090 GPU processor.}
\label{tab:compute_table}
\end{table}

\begin{figure}[t]
\begin{center}
\includegraphics[width=1\linewidth]{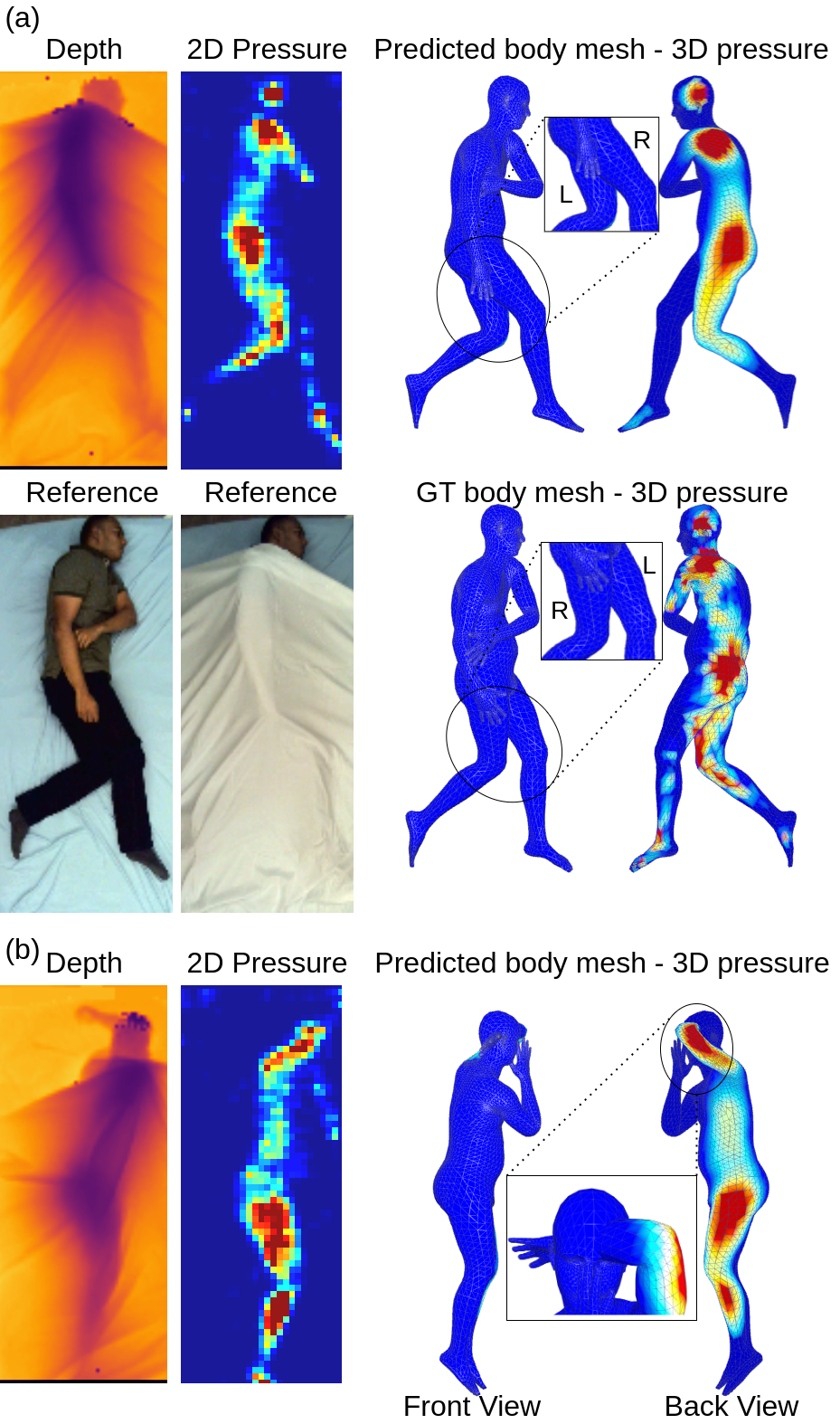}
\end{center}
   \caption{Failure cases of BodyMAP. (a) The Model predicts flipped orientation for legs stemming from inherent body symmetries. (b) Predicted mesh interpenetration failure example.}
\label{fig:s_fail}
\end{figure}

\section{Future Work Opportunities}
\label{sec:supp_future}

While BodyMAP demonstrates proficient estimation of body mesh and 3D applied pressure map, its translation into clinical efficacy requires validation through real-world studies. Clinical settings inherently differ from training scenarios, encompassing variations in mattresses, diverse body poses, partially elevated or tilted mattresses for actuated hospital beds, and the presence of additional objects like pillows and medical instruments. It is also crucial to quantify the benefits derived from the body mesh and 3D pressure map predictions \& visualizations in improving pressure injury prevention in real clinical scenarios.

Additionally, our method encounters occasional false positives. For instance, in \cref{fig:s_fail}(a), the misalignment of leg orientation, attributed to inherent body symmetries, leads to an inaccurate 3D pressure map. Future enhancements may stem from expanded training datasets.

Furthermore, instances of mesh inter-penetrations can occur in our predictions. As depicted in \cref{fig:s_fail}(b), the left arm penetrates the head. Although incorporating loss terms to mitigate such occurrences is viable, our method currently leverages a fixed open-hand pose similar to BPW~\cite{clever2022bodypressure}, contributing to occasional unnatural orientations. 

Similar to BPW~\cite{clever2022bodypressure}, our current approach fails to account for pressure from self-contact. Integrating methods that consider physics behind self-contact~\cite{hassan2019resolving, muller2021self, fieraru2021learning} could potentially augment our method's performance.

Our proposed BodyMAP-WS model employs an implicit learning strategy, effectively eliminating the need for ground truth 3D pressure map data. Although BodyMAP-WS requires a pre-trained mesh model, which traditionally necessitates 3D ground truth mesh data, our methodology paves the way for a fully self-supervised model. In this paradigm, both the body mesh and 3D applied pressure map could be learned without relying on any 3D labels. This advancement holds promise for the field, as it opens avenues to truly scale up to large-scale unlabeled data collected through real-world systems.

The optimized architecture of BodyMAP has significantly reduced inference time, enabling real-time analysis. This advancement creates opportunities to train models for predicting body mesh and 3D applied pressure maps on video data. This expanded capability would allow to capture richer insights and further improve the pressure ulcer prevention process.

\section{Video Results}
\label{sec:supp_video}

We release video results on our project website, visualizing the body mesh with a 3D pressure map to compare our method with ground truth and BPW~\cite{clever2022bodypressure}. The videos enable detailed visual analysis of the predicted pressure along with the predicted body mesh. 

\begin{figure*}
\begin{center}
\includegraphics[width=1.0\linewidth]{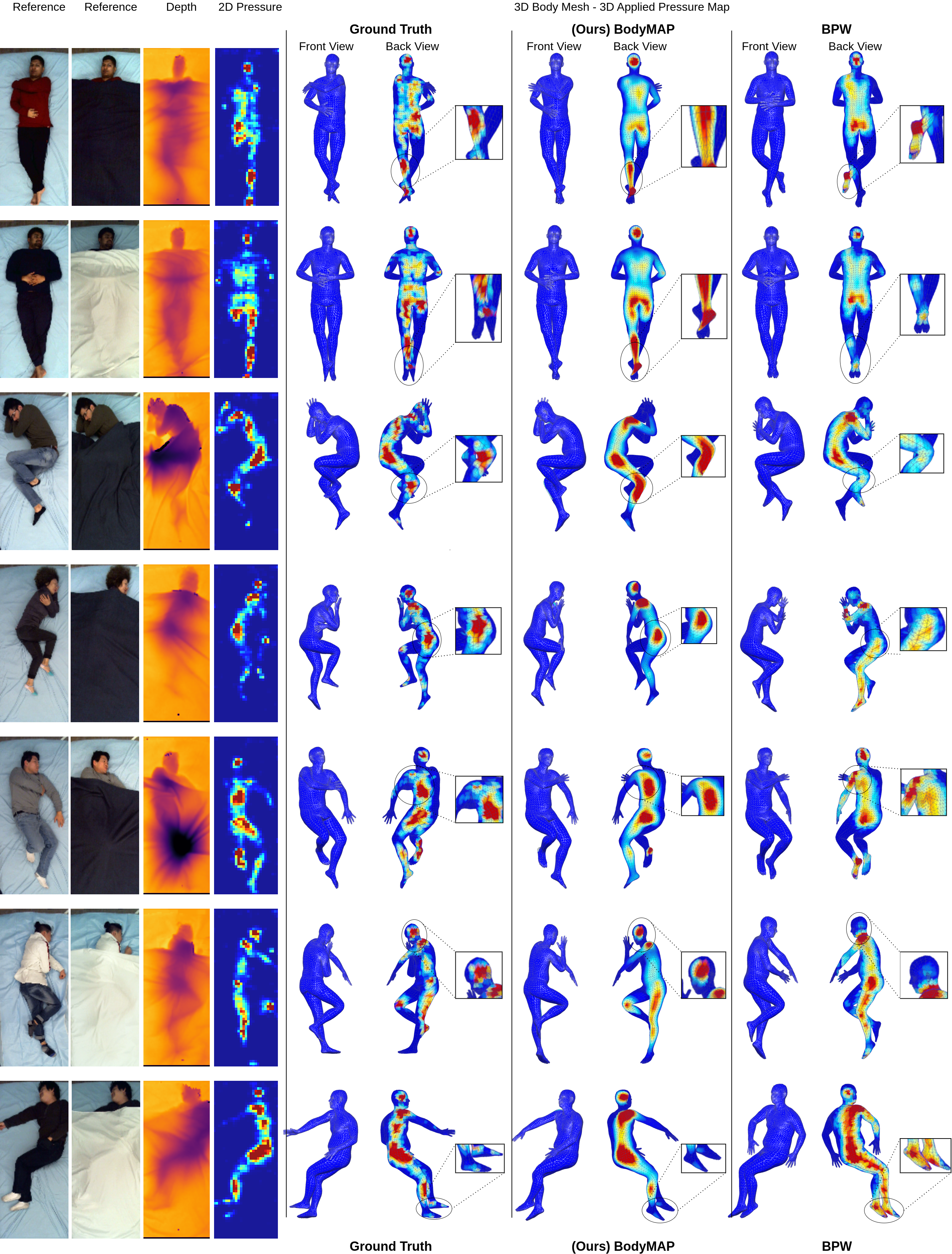}
\end{center}
   \caption{Results of inferring body mesh and 3D applied pressure map from examples in the SLP~\cite{liu2019seeing, liu2022simultaneously} test set.}
\label{fig:s_c1}
\end{figure*}

\begin{figure*}
\begin{center}
\includegraphics[width=1.0\linewidth, trim={0cm, 18cm, 0cm, 0cm}]{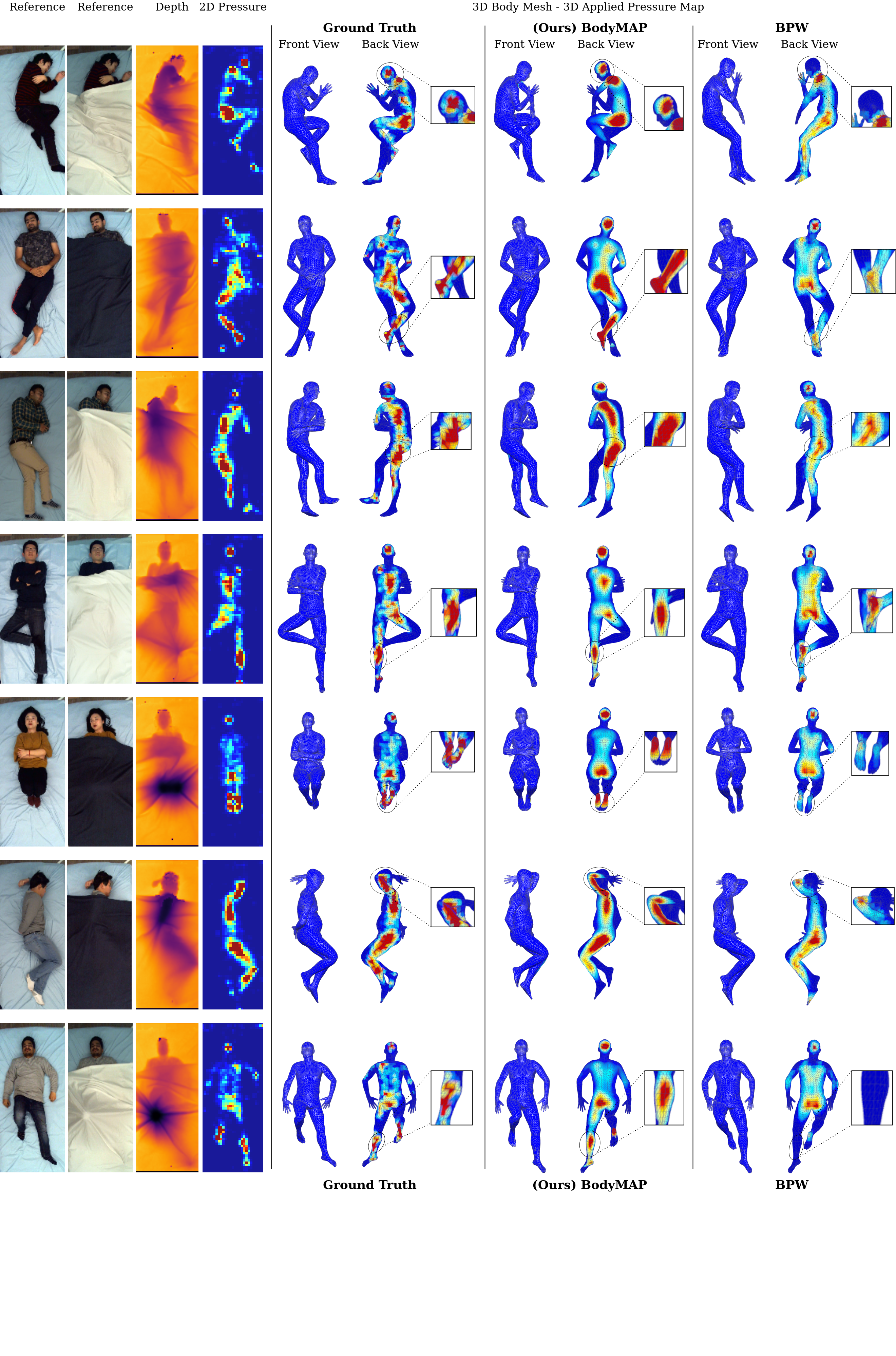}
\end{center}
   \caption{More results of inferring body mesh and 3D applied pressure map from examples in the SLP~\cite{liu2019seeing, liu2022simultaneously} test set.}
\label{fig:s_c2}
\end{figure*}

Additionally, we provide more visual results comparing our method with ground truth and BPW~\cite{clever2022bodypressure} in \cref{fig:s_c1} \& \cref{fig:s_c2}.

\end{document}